\documentclass{article}

\usepackage{float}
\usepackage{xcolor}

\newcommand{\ghurl}{https://github.com/AmitoVrito/Traceprop}

\usepackage{amsmath,amssymb,amsthm}
\usepackage{booktabs}
\usepackage{graphicx}
\newcommand{\fittab}[1]{\resizebox{\ifdim\width>\columnwidth\columnwidth\else\width\fi}{!}{#1}}
\usepackage{pgfplots}
\pgfplotsset{compat=1.18}
\usepackage{tikz}
\usetikzlibrary{arrows.meta,positioning,shapes.geometric,fit,backgrounds,calc}
\usepackage{listings}
\usepackage{microtype}
\usepackage{enumitem}
\usepackage{caption}
\usepackage{subcaption}
\usepackage{url}
\usepackage{xurl}
\usepackage{multirow}
\usepackage{array}
\usepackage{makecell}
\usepackage{dblfloatfix}

\usepackage{hyperref}

\usepackage[accepted]{mlsys2025}
\makeatletter\renewcommand{\Notice@String}{}\makeatother

\definecolor{codebg}{rgb}{0.97,0.97,0.97}
\definecolor{codegreen}{rgb}{0.13,0.54,0.13}
\definecolor{codepurple}{rgb}{0.58,0,0.82}
\definecolor{codeblue}{rgb}{0.0,0.27,0.65}
\lstdefinestyle{python}{
  backgroundcolor=\color{codebg},
  commentstyle=\color{codegreen}\itshape,
  keywordstyle=\color{codeblue}\bfseries,
  stringstyle=\color{codepurple},
  basicstyle=\ttfamily\footnotesize,
  breakatwhitespace=false,
  breaklines=true,
  captionpos=b,
  keepspaces=true,
  language=Python,
  showspaces=false,
  showstringspaces=false,
  showtabs=false,
  tabsize=2,
  frame=single,
  rulecolor=\color{gray!40},
  xleftmargin=4pt,
  xrightmargin=4pt,
}

\newtheorem{theorem}{Theorem}

\newtheorem{corollary}{Corollary}

\hypersetup{
  colorlinks=true,
  linkcolor=blue!70!black,
  citecolor=green!50!black,
  urlcolor=blue!70!black,
  pdfauthor={Amit Nautiyal},
}

\mlsystitlerunning{Traceprop: Training Data Attribution in a Single Pass}

\begin{document}

\twocolumn[
\mlsystitle{Traceprop: Training Data Attribution in a Single Pass}

\begin{mlsysauthorlist}
\mlsysauthor{Amit Nautiyal}{ind}
\end{mlsysauthorlist}

\mlsysaffiliation{ind}{Independent Researcher}
\mlsyscorrespondingauthor{Amit Nautiyal}{research.amit.n@gmail.com}

\mlsyskeywords{data attribution, machine learning systems, influence functions, provenance}

\vskip 0.3in

\begin{abstract}
Data attribution tools used in practice, TRAK, LoGRA/LogIX, and EK-FAC, run after training:
they recompute per-sample gradients in a separate pass over the training set, and
curvature-aware variants need a further pass to estimate covariance. Traceprop records
projected per-sample gradients and K-FAC covariance statistics inside the training backward
pass itself. On Pythia-1B LoRA fine-tunes on an A100, this adds $2.6\%$ to training time,
against $10.7\%$ for LogIX's best inline configuration, random-init ($4.1\times$ lower,
one-sided Mann--Whitney $p{=}9\times10^{-5}$, $n{=}10$ per arm); at Pythia-6.9B the numbers are
$11.3\%$ and $27.0\%$ ($p{=}0.004$). LogIX's default configuration, PCA-init, is both slower
and lower quality than random-init, so we compare against random-init throughout. On a small
transformer where LDS is measurable, Traceprop matches LogIX's best configuration: pooled LDS
difference $+0.0022$, $95\%$ CI ${[}-0.0038,+0.0075{]}$. Two things do not work: at
Pythia-160M/SST-2, every method is indistinguishable from a low noise ceiling, and on
planted-backdoor and mislabel detection, gradient attribution does not beat gradient norm or
representation similarity. The contribution here is systems, not a new estimator: LogIX's
attribution quality, in one pass instead of two.
\end{abstract}
]

\printAffiliationsAndNotice{}

\hypersetup{pdfauthor={Amit Nautiyal}}

\section{Introduction}
\label{sec:intro}

Data attribution, the problem of deciding which training examples were responsible for a
given prediction, now underlies a range of tasks: debugging, data valuation, curation,
copyright and privacy analysis, and machine unlearning. The methods that dominate in
practice (TRAK \citep{park2023trak}, LoGRA/LogIX \citep{logix2025}, EK-FAC
\citep{grosse2023studying}, DataInf \citep{kwon2024datainf}) all share the same shape:
they run after training. Given a finished checkpoint, each of them makes a dedicated pass
over the whole training set, computing and projecting every example's gradient to build
the store that later queries read from; the estimator that turns a raw gradient dot product
into strong attribution, K-FAC-style preconditioning, needs a further separate covariance
pass on top of that. Getting ready to attribute a model's predictions therefore costs one or
two extra traversals of the training set, often more than the fine-tune itself.

Both of those extra passes are avoidable. The training loop already computes a per-sample
gradient on its own backward pass and then throws it away; the covariance a preconditioner
needs can be built from the same per-step quantities. Traceprop keeps them instead.
At each step, forward and backward hooks on the tracked linear layers reconstruct the
per-sample gradient through the factored identity $\partial\mathcal{L}/\partial W = \sum_t
g_t \otimes a_t$. We sketch it with small Kronecker-factored projections held on the
accelerator (\S\ref{sec:factored}) and accumulate the sketch-space covariance from those same
projected factors (\S\ref{sec:kfac}), all in the one pass training already runs. Only the
compressed sketch and its covariance ever leave the GPU. The result costs a quarter of the
overhead of the best existing tool at 1B parameters and under half at 6.9B, even when that
tool also runs inline (\S\ref{sec:overhead}).

\paragraph{Why this is a systems contribution.} The saving lands on a cost practitioners
already pay. A comparably-capable post-hoc pipeline spends a full extra pass for gradients
and, for K-FAC-quality scores, a second full pass for covariance; inline logging spends both
inside the training step already running. This holds at two LLM scales on A100 GPUs, and the
mechanism is a context manager that leaves the model and optimizer untouched. Every stored
gradient also carries its source-file lineage, so the same pass answers audit-trail queries
under EU AI Act Article~26 (Appendix~\ref{app:beyond}).

\paragraph{Contributions.}
\begin{enumerate}[leftmargin=*, label=(\arabic*), topsep=2pt, itemsep=1pt]
  \item \textbf{A single-pass mechanism}: inline per-sample gradient logging via factored
  last-layer hooks and on-accelerator sparse-JL sketching, plus inline K-FAC preconditioning
  that accumulates sketch-space covariance from the same projected factors the sketch already
  computes, with no second covariance pass. Query-time whitening is unit-tested equivalent to
  an independent separate-pass computation. (\S\ref{sec:design})
  \item \textbf{A $\sim$4$\times$ overhead result at equal quality}: on Pythia-1B (A100),
  Traceprop's total single-pass overhead is $2.6\%$ against LogIX's best single-pass
  configuration (random-init) at $10.7\%$ ($p{=}9\times10^{-5}$, $n{=}10$); the pattern
  holds at Pythia-6.9B ($11.3\%$ vs.\ $27.0\%$, $p{=}0.004$, $n{=}5$). LogIX's two-pass
  PCA-init variant is also worse on quality (below). A dense, non-preconditioned logging-only
  path stays at $0.30$--$1.08\%$ across four smaller models; that number is a labeled component
  of the design space, distinct from the headline single-pass result above.
  (\S\ref{sec:overhead}, Figure~\ref{fig:overhead_bars})
  \item \textbf{Quality parity with LogIX's true best}: on a from-scratch LoRA transformer,
  Traceprop ties LogIX's random-init preconditioning, which is both the cheapest and the
  strongest configuration LogIX has. LogIX's PCA-init variant is worse on quality and needs a
  second pass: worse on both axes. (\S\ref{sec:quality})
  \item \textbf{Negative results}: at transformer scale (Pythia-160M/SST-2), all
  methods are statistically indistinguishable against a low noise ceiling; a planted-backdoor
  and mislabel-detection probe shows no attribution-specific advantage over free baselines on
  either task. (\S\ref{sec:negative})
  \item Applications carried by the same inline provenance, evaluated at data-systems scale
  in Appendix~\ref{app:beyond}: computation-level lineage, source-stratified attribution (SAB
  v1 benchmark), provenance-guided unlearning, and an EU AI Act Article~26 export schema.
\end{enumerate}

Code:\par\noindent{\scriptsize\mbox{\url{\ghurl}}}

\newcommand{\mlinspect}{\textsc{mlinspect}}
\newcommand{\traceprop}{Traceprop}
\newcommand{\trak}{TRAK}

\section{Background}
\label{sec:background}

\paragraph{Training data attribution.}
\citet{koh2017understanding} introduced influence functions:
\begin{equation}
  \mathcal{I}(z_i, z_\text{test}) \approx -\nabla_\theta \mathcal{L}(z_\text{test})^\top H_\theta^{-1} \nabla_\theta \mathcal{L}(z_i).
  \label{eq:influence}
\end{equation}
TRAK \citep{park2023trak} approximates Eq.~\eqref{eq:influence} via random JL projection and a
per-class linearized model; a 5-checkpoint TRAK ensemble in our setup reaches LDS $=0.0290$ on
CIFAR-2/ResNet-9, taking $691\,\text{s}$ on GPU. DSDM \citep{engstrom2024dsdm} uses datamodel
regression for dataset distillation. dattri \citep{deng2024dattri}, DataInf
\citep{kwon2024datainf}, LESS \citep{xia2024less}, TrackStar \citep{chang2024trackstar}, and
Kronfluence \citep{kronfluence2024} provide further attribution implementations; the last two
are built around the same K-FAC-style covariance statistics LogIX uses. LogIX preconditions
with K-FAC \citep{grosse2023studying}-style covariance, computed, as we confirm in
\S\ref{sec:kfac}, in the same kind of low-dimensional projected space Traceprop uses. None of
these tools connects to data lineage.

\paragraph{Provenance and compliance context.}
Database provenance has a canonical taxonomy \citep{cheney2009provenance} and classical
relational systems (Perm \citep{glavic2009perm}, Smoke \citep{psallidas2018smoke}, GProM
\citep{feng2017gprom}, ProvDB \citep{miao2017provdb}), but these operate at the SQL tuple or
relational-operator level and do not extend to tensor computations. ML pipeline-provenance tools
(MLflow \citep{mlflow2018}, DVC \citep{dvc2020}, \mlinspect\ \citep{grafberger2021mlinspect},
DSLog \citep{namaki2025dslog}) track artifacts or preprocessing lineage but provide no
attribution scores. Machine unlearning has moved from exact \citep{bourtoule2021machine} to
certified-approximate \citep{sekhari2021remember} and efficient gradient-correction
\citep{warnecke2021machine,yao2023llmunlearn} methods, all treating the forget set as externally
specified, unlike Traceprop's provenance-derived forget sets. EU AI Act Article~26 requires
high-risk AI deployers to maintain automatically generated audit logs
\citep{euaiact2024,epimco2026}, an obligation neural-network memorization makes technically
non-trivial \citep{carlini2022secret,feldman2020neural}.

\section{Design}
\label{sec:design}

\begin{figure}[t]
\centering
\begin{tikzpicture}[
  box/.style={draw=black, thin, rounded corners=1pt, minimum width=2.6cm, minimum height=0.5cm,
              align=center, font=\scriptsize, fill=white},
  emph/.style={draw=black, thick, rounded corners=1pt, minimum width=2.6cm,
               minimum height=0.5cm, align=center, font=\scriptsize, fill=gray!12},
  arrow/.style={-stealth, thin, color=black},
]
\node[box]  (fwd)  at (0,1.1)    {Forward hook\\caches $a$};
\node[box]  (bwd)  at (2.8,1.1)  {Backward hook\\caches $g$};
\node[emph] (sk)   at (1.4,0)    {Factored sketch\\$(Pg_t)\otimes(Qa_t)$};
\node[box]  (buf)  at (0,-1.1)   {Sketch buffer\\(device)};
\node[box]  (cov)  at (2.8,-1.1) {Inline covariance\\$\hat G,\hat A$};
\node[emph] (host) at (1.4,-2.2) {Host drain\\(once per run)};
\draw[arrow] (fwd) -- (sk);
\draw[arrow] (bwd) -- (sk);
\draw[arrow] (sk) -- (buf);
\draw[arrow] (sk) -- (cov);
\draw[arrow] (buf) -- (host);
\draw[arrow] (cov) -- (host);
\end{tikzpicture}
\caption{Everything inside one backward pass: hooks $\to$ factored sketch
(\S\ref{sec:factored}) $\to$ device-side sketch buffer and inline covariance
(\S\ref{sec:kfac}) $\to$ a single host transfer at the end of the run.}
\label{fig:hooks_flow}
\end{figure}

\subsection{Inline per-sample logging}
\label{sec:inline}
For an adapted linear layer $y = xW^\top$ with input activations $a\in\mathbb{R}^{B\times T\times d_\text{in}}$
and output gradients $g\in\mathbb{R}^{B\times T\times d_\text{out}}$, the per-sample gradient of the
scalar loss is $\sum_t g_{b,t}\otimes a_{b,t}$. A forward hook caches $a$ and a full-backward
hook caches $g$; at each optimizer step we form the per-sample outer products only for the
tracked (small, low-rank LoRA) layers and sketch them (\S\ref{sec:factored}) into a small
device-side buffer, drained to host once per run. No dense per-sample gradient and no large
projection intermediate are ever materialized off-accelerator. Because a single batched
backward carries per-sample output gradients unreduced, the hook-computed per-sample gradients
equal those obtained by backpropagating each example individually; we verify cosine similarity
$=1.000000$ against per-example autograd for every sample in a batch (a global scale from the
loss-reduction convention cancels in dot-product ranking). This per-example path is
implemented by a dedicated \texttt{LoRAGradientLogger}, distinct from the generic
\texttt{TrainingContext} described in Appendix~\ref{app:beyond} (which records batch-mean
gradients for models without a LoRA adapter).

\subsection{Kronecker-factored sketch}
\label{sec:factored}
A dense sparse-JL projection of the per-sample gradient, $\tilde g_i = Jg_i$, uses a sparse
Johnson--Lindenstrauss matrix $J\in\mathbb{R}^{k\times p}$ whose entries are drawn i.i.d.\ as
$\pm\sqrt{3/k}$ each with probability $1/6$ and $0$ with probability $2/3$ (see
Eq.~\eqref{eq:jl} for the full derivation). This does not scale to tracking every layer:
its projection matrix is $p\times k$ over the full flattened gradient dimension $p =
\sum_\text{layer} d_\text{out}d_\text{in}$, tens of gigabytes at all-layer scope on a 1B-parameter
model. We instead sketch the Kronecker structure directly: $\sum_t g_t\otimes a_t \approx
\sum_t (Pg_t)\otimes(Qa_t)$ with small sparse-JL maps $P\in\mathbb{R}^{k_\text{out}\times
d_\text{out}}$, $Q\in\mathbb{R}^{k_\text{in}\times d_\text{in}}$ applied per module, never
materializing the $d_\text{out}\cdot d_\text{in}$ outer product or a large projection matrix.
The dot product of two such sketches is an unbiased estimate of the true gradient dot product,
which is what attribution needs. This factored path is what scales to all-layer tracking where
the dense path cannot run at all (\S\ref{sec:overhead}).

\subsection{Inline K-FAC preconditioning}
\label{sec:kfac}
A raw sketch dot product is a weaker attribution estimator than one preconditioned by the loss
curvature; LogIX's own gain from K-FAC-style preconditioning is real (\S\ref{sec:quality}), but
its implementation accumulates that covariance in a dedicated pass separate from logging.
Traceprop accumulates it inline, reusing the projected factors the sketch already computes.
Per module $j$, over the same backward pass that produces the sketch, we accumulate
$\hat G_j = \sum_t (Pg_t)(Pg_t)^\top$ and $\hat A_j = \sum_t (Qa_t)(Qa_t)^\top$, two small
$k\times k$ matrices built from quantities already in registers for the sketch, at the marginal
cost of two small matmuls per step and no extra pass over the data. At query time, each stored
per-module sketch block $S_j$ (shape $k_\text{out}\times k_\text{in}$) is whitened as
$W_{G_j} S_j W_{A_j}$, with $W_{G_j} = (\hat G_j + \lambda_{G_j}I)^{-1/2}$ from a damped
symmetric eigendecomposition and $\lambda_{G_j} = \delta\cdot\mathrm{tr}(\hat G_j)/k$ scaled by
the module's own mean eigenvalue so the damping is invariant to the covariance's arbitrary
overall scale; $\delta$ is chosen per module on a held-out validation split, kept separate from
the evaluation split we report results on. The Frobenius inner product of two whitened sketches
then equals $\mathrm{tr}(S_i^\top \hat G_j^{-1} S_j \hat A_j^{-1})$, the K-FAC-preconditioned
influence, computed entirely in the low-dimensional sketch space. Unit tests confirm the
inline-accumulated covariance matches an independent separate-pass computation at
$\mathrm{rtol}=10^{-6}$, and that the whitened dot product matches a direct computation of the
trace expression at $10^{-8}$ (Appendix~\ref{app:tests}). These are the tests the
``one pass, not two'' claim rests on.

\section{Implementation}
\label{sec:impl}

\texttt{LoRAGradientLogger} is a context manager: entering it registers forward and full-backward
hooks on the tracked \texttt{nn.Linear} modules (PEFT-wrapped or plain); exiting it removes the
hooks and drains any buffered sketches to host. Applying it to PEFT-wrapped LoRA adapters needed
one compatibility fix, a read-only property forwarding PEFT's direct weight-tensor read. Sketches
are accumulated in a device-side buffer (\texttt{buffer=True}) so the host transfer, and the
synchronization it forces, happens once at the end of a run rather than every step, keeping the
projection off the training loop's critical path. The projected result and, when
\texttt{inline\_precond=True}, the sketch-space covariances are the only tensors ever copied off
the accelerator. Nothing about the optimizer, the model's forward pass, or its gradients is
altered; adding or removing the logger changes only what is additionally recorded.

\section{Evaluation}
\label{sec:eval}

\subsection{Overhead}
\label{sec:overhead}
\paragraph{Single-pass headline.} Table~\ref{tab:overhead_headline} reports the main
comparison: matched-repeat, Mann--Whitney-tested overhead of Traceprop's full single pass
(factored logging \emph{and} inline K-FAC covariance) against LogIX at its two possible
configurations, on identical Pythia-1B LoRA fine-tunes (NVIDIA A100, $n{=}10$ repeats each arm,
$200$ steps, batch $8$, sequence $128$, all-layer tracking, $34$ tracked modules, storage-matched
$k{=}8$). LogIX's PCA-init variant, its default, needs a separate full-dimensional covariance
pass \emph{before} the LoRA adapter it logs through even exists. A sanity check, the ratio of
covariance-pass time to one plain forward--backward, comes out at $1.38\times$, which is what
full-dimensional accumulation should cost, so this is real work and not a timing artifact:
$147.9\%$ total. Getting a fair baseline took one fix beyond that: LogIX's \texttt{watch()}
freezes parameters outside the tracked scope to save compute and is called a second time
internally by \texttt{add\_lora()}, which silently gave a naively-timed LogIX model less
training work than the plain baseline. We restore \texttt{requires\_grad} after every
\texttt{watch()} call in both arms. LogIX also supports a random-init variant that gathers
covariance and logs gradients in the same pass (confirmed by inspecting
\texttt{LogOption.setup}, which accepts forward, backward, and grad plugins together), making
it single-pass like Traceprop. At $10.7\%$ it is LogIX's fastest configuration, and
\S\ref{sec:quality} shows it is also LogIX's best on quality, so PCA-init is worse on both axes
and appears here only for completeness. Against LogIX's true best, Traceprop's $2.6\%$ is a
$4.1\times$ reduction, significant at $p{=}9\times10^{-5}$ (one-sided Mann--Whitney, LogIX-random
overhead-\% samples $>$ Traceprop overhead-\% samples, matched $n{=}10$).

\begin{table}[t]\centering\small
\setlength{\tabcolsep}{4pt}
\fittab{%
\begin{tabular}{lccc}
\toprule
Config & Passes & Overhead (\%) & vs.\ Traceprop \\
\midrule
LogIX (PCA-init)    & 2 & $147.914$ & dominated \\
LogIX (random-init) & 1 & $10.659$  & $4.1\times$ ($p{=}9\text{e-}5$) \\
\textbf{Traceprop}   & \textbf{1} & \textbf{2.588} & --- \\
\bottomrule
\end{tabular}}
\caption{Attribution-readiness overhead, Pythia-1B, NVIDIA A100, all-layer tracking, $k{=}8$,
matched $n{=}10$ repeats. Traceprop's $2.588\%$ splits as $0.855\%$ logging $+1.703\%$ inline
covariance. LogIX-PCA's $147.914\%$ splits as $9.732\%$ logging $+138.183\%$ covariance pass
(sanity: covariance-pass / plain-forward-backward $= 1.38\times$). LogIX-random needs one pass
but is not LogIX's default configuration.}
\label{tab:overhead_headline}
\end{table}

\begin{figure}[t]\centering
\begin{tikzpicture}
\begin{axis}[
  ybar, bar width=16pt, width=0.85\linewidth, height=4.4cm,
  enlarge x limits=0.35,
  symbolic x coords={1B,6.9B}, xtick=data,
  xticklabels={1B (fp32),6.9B (bf16)},
  x tick label style={font=\scriptsize}, y tick label style={font=\scriptsize},
  ylabel={Overhead vs.\ plain training (\%)}, ylabel style={font=\scriptsize},
  ymin=0, ymax=32,
  legend columns=-1,
  legend style={font=\scriptsize, draw=none, at={(0.5,1.05)}, anchor=south},
  axis line style={line width=0.6pt}, tick style={line width=0.6pt},
  error bars/y dir=both,
  error bars/y explicit,
  error bars/error bar style={line width=0.8pt, black},
  error bars/error mark options={mark size=3pt, line width=0.8pt, black},
]
\addplot+[fill=gray!30, draw=black, line width=0.6pt] coordinates
  {(1B,2.588) +- (0,0.17) (6.9B,11.274) +- (0,1.41)};
\addplot+[fill=gray!75, draw=black, line width=0.6pt] coordinates
  {(1B,10.659) +- (0,0.06) (6.9B,26.987) +- (0,0.33)};
\legend{Traceprop,LogIX-random}
\node[font=\small, anchor=south, xshift=-8pt] at (axis cs:1B,2.988) {2.6};
\node[font=\small, anchor=south, xshift=-8pt] at (axis cs:6.9B,12.984) {11.3};
\node[font=\small, anchor=south, xshift=8pt] at (axis cs:1B,10.919) {10.7};
\node[font=\small, anchor=south, xshift=8pt] at (axis cs:6.9B,27.517) {27.0};
\end{axis}
\end{tikzpicture}
\caption{Overhead vs.\ plain training, Pythia-1B (fp32) and Pythia-6.9B (bf16), NVIDIA A100.
Bars show the median (matching Table~\ref{tab:overhead_headline} and the text, labeled above
each bar); error bars show $\pm1$ sample std across the $n{=}10$ (1B) / $n{=}5$ (6.9B) repeats.
A single warm-up-adjacent outlier at 6.9B widens Traceprop's error bar there
(Appendix~\ref{app:overhead_investigation}).}
\label{fig:overhead_bars}
\end{figure}

\paragraph{7B, a different dtype.} The same measurement at Pythia-6.9B (bf16, single NVIDIA
A100, $n{=}5$ repeats each arm, $66$ tracked modules, $k{=}8$, LogIX-PCA arm skipped since its
full-dimensional covariance pass is already prohibitive at 1B and is dominated on quality
regardless) gives Traceprop $11.3\%$ versus LogIX-random $27.0\%$, significant at
$p{=}0.004$ ($n{=}5$ both arms): Traceprop's margin over LogIX-random is $4.1\times$ at 1B and
$2.4\times$ at 6.9B, each ratio computed separately at its own model size and dtype
(Figure~\ref{fig:overhead_bars}). We avoid reading $2.6\%\to11.3\%$ as a scaling trend, because
the 1B measurement runs in fp32 and the 7B measurement runs in bf16 (auto-selected for memory),
which changes tensor-core code paths on top of changing model size; the two headline numbers
are each reported with their own dtype rather than plotted as one curve. Both models' overhead
grew relative to their own single-pass baseline going from 1B to 7B. We investigated one
candidate implementation cause and ruled it out by direct measurement (Appendix~\ref{app:overhead_investigation}); the growth itself is an open question
(\S\ref{sec:limitations}).

\paragraph{Dense logging-only component.} Table~\ref{tab:inline_overhead} reports the
non-preconditioned, last-block-only dense path in isolation: interleaved, repeated ($\times20$)
measurements of a LoRA fine-tune step with and without inline logging (no covariance
accumulation), across four models on an L4. This is the cheapest configuration Traceprop
supports and the one earlier drafts of this work headlined; it remains a valid, useful
data point, $0.30$--$1.08\%$ throughput overhead with a $2\,\text{KB}$-per-example store. It
omits preconditioning and cannot scale past last-block tracking (\S\ref{sec:factored}), so we
report it here as a labeled component of the design space rather than folding it into the main
overhead claim.

\begin{table}[t]\centering\small
\setlength{\tabcolsep}{4pt}
\fittab{%
\begin{tabular}{lcccc}
\toprule
Model & base step (ms) & dtype & throughput \% & synced bound \% \\
\midrule
GPT-2 (124M)     & 96.4  & fp32 & $1.08\pm0.14$ & $1.67\pm0.32$ \\
Pythia-410M      & 208.0 & fp32 & $0.30\pm0.16$ & $0.39\pm0.17$ \\
Pythia-1B        & 424.0 & fp32 & $0.32\pm0.11$ & $0.36\pm0.12$ \\
Pythia-2.8B      & 332.6 & bf16 & $0.33\pm0.11$ & $0.11\pm0.17$ \\
\bottomrule
\end{tabular}}
\caption{Dense-path, logging-only overhead component (L4, last-block, sparse-JL projection,
$k=512$, 2\,KB/example store, $20$ interleaved repeats, no preconditioning). Throughput overhead
(realistic; the projection overlaps compute) stays under $1.1\%$ on every model. Pythia-2.8B runs
in bf16 to fit LoRA fine-tuning on a 24GB GPU; the other three run in fp32. This is a labeled
component of the design space (\S\ref{sec:overhead}); the paper's headline overhead number is
Table~\ref{tab:overhead_headline}.}
\label{tab:inline_overhead}
\end{table}

\paragraph{2-GPU DDP.} A structural check that the logger adds no cross-rank communication
(2-way \texttt{DistributedDataParallel}, dense logging-only path) found no measurable overhead
at 2 ranks ($0.372\pm2.99\%$ on both ranks, noise at this timing floor); full setup and
caveats in Appendix~\ref{app:supplementary}.

\paragraph{Head-to-head vs.\ post-hoc extraction.} A supplementary, timing-only comparison
against a full post-hoc extraction pass (dense projection, last-block scope, unaffected by the
LogIX-specific fixes above) gives $60$ to $242\times$ against a single post-hoc checkpoint and
$301$ to $1211\times$ against a five-checkpoint TRAK ensemble, growing with model size; full
table in Appendix~\ref{app:headtohead}. This isolates the projection/write cost
alone, separate from the full single-pass number reported above; it shows a structural gap
(post-hoc pays a full pass, times the ensemble size, while inline pays a fraction of one step)
that holds regardless of the preconditioning result in \S\ref{sec:kfac}.

\subsection{Attribution quality}
\label{sec:quality}
LDS needs hundreds of retraining runs. Within our compute budget that is tractable only on a
small from-scratch LoRA transformer; the 1B/6.9B models used for the overhead numbers are too
large to retrain hundreds of times. We measure it with the Linear
Datamodeling Score (LDS \citep{park2023trak}: retraining on random data subsets and correlating
$\text{mask}\cdot\text{scores}$ with the retrained test margins), all-layer scope, storage-matched
to LogIX ($k{=}8$), pooled over 5 target-model seeds with a hierarchical bootstrap (resample
seeds, then subsets and held-out evaluation examples within seed) and a held-out validation split
used only to choose damping. Table~\ref{tab:quality_absolute} gives absolute LDS for every method
under both LogIX initialization strategies.

\begin{table}[t]\centering\small
\setlength{\tabcolsep}{3pt}
\fittab{%
\begin{tabular}{lcc}
\toprule
Method & LogIX random-init & LogIX PCA-init \\
\midrule
LogIX dot                    & $0.588\pm0.081$ & $0.579\pm0.072$ \\
LogIX preconditioned         & $0.633\pm0.078$ & $0.593\pm0.076$ \\
LogIX preconditioned, tuned  & $0.639\pm0.078$ & $0.629\pm0.080$ \\
Traceprop dense dot          & $0.597\pm0.072$ & --- \\
Traceprop factored dot       & $0.586\pm0.071$ & --- \\
\textbf{Traceprop factored + inline precond.} & $\mathbf{0.641\pm0.081}$ & --- \\
Random baseline              & \multicolumn{2}{c}{$0.001\pm0.009$} \\
\bottomrule
\end{tabular}}
\caption{Absolute LDS (mean $\pm$ std over 5 seeds, held-out eval split, all-layer, $k{=}8$).
Traceprop's factored sketch is only evaluated under random-init masks/margins (it does not
depend on LogIX's LoRA initialization); the PCA-init column exists to compare LogIX against
itself under the two initializations available to it.}
\label{tab:quality_absolute}
\end{table}

\begin{figure}[t]\centering
\begin{tikzpicture}
\begin{axis}[
  width=\linewidth, height=4.4cm, xmode=log,
  xlabel={1B overhead (\%, log scale)}, ylabel={Tiny-model LDS (tuned, preconditioned)},
  xlabel style={font=\scriptsize}, ylabel style={font=\scriptsize},
  x tick label style={font=\scriptsize}, y tick label style={font=\scriptsize},
  xmin=1.5, xmax=250, ymin=0.60, ymax=0.66,
  only marks, mark size=2.2pt,
]
\addplot+[mark=*, black, fill=black] coordinates {(2.588,0.641)}
  node[above right, font=\tiny] {TraceProp};
\addplot+[mark=square*, black, fill=gray!45] coordinates {(10.659,0.639)}
  node[above right, font=\tiny] {LogIX random-init};
\addplot+[mark=triangle*, black, fill=white] coordinates {(147.914,0.629)}
  node[above left, font=\tiny] {LogIX PCA-init};
\end{axis}
\end{tikzpicture}
\caption{Overhead vs.\ quality across the three configurations. \emph{Different setups on each
axis}: $x$ is the 1B, A100 overhead measurement of \S\ref{sec:overhead}
(Table~\ref{tab:overhead_headline}); $y$ is the tiny from-scratch-transformer LDS of this
section (Table~\ref{tab:quality_absolute}, preconditioned-and-tuned row, each method's own
init). These are two separate measurements plotted on shared axes to show the trade-off, not
one paired experiment. TraceProp sits up and to the left (lower overhead, no quality loss);
LogIX-PCA is dominated on both axes. TraceProp and LogIX-random are
statistically tied (paired $95\%$ CI includes zero, \S\ref{sec:quality}).}
\label{fig:quality_overhead_scatter}
\end{figure}

\paragraph{In our setting, LogIX's default configuration is dominated.} A pooled hierarchical
bootstrap directly comparing LogIX's two initializations, both damping-tuned on the same
held-out split and evaluated on identical retraining subsets (masks and margins verified
identical before comparing), gives random-init $-$ PCA-init $= +0.0099$, $95\%$ CI
$[+0.0053,+0.0145]$, $100\%$ of bootstrap resamples positive. Combined with
\S\ref{sec:overhead}'s overhead result (random-init is the only single-pass LogIX
configuration), PCA-init, LogIX's default, is strictly dominated here: worse quality and more
compute.

\paragraph{Traceprop ties LogIX's actual best.} Against LogIX's random-init, tuned,
preconditioned estimator, the pooled bootstrap gives Traceprop (factored $+$ inline
preconditioning) $-$ LogIX-random $= +0.0022$, CI $[-0.0038,+0.0075]$ at $k{=}8$; $-0.0009$, CI
$[-0.0075,+0.0053]$ at $k{=}7$; both CIs include zero. This is parity: Traceprop matches the
strongest attribution quality either tool reaches at this scale, in a single pass, while LogIX
needs a second pass to reach any preconditioned quality at all (PCA-init) or ties Traceprop
while also being single-pass (random-init). Preconditioning itself is a clear within-method win
over the unpreconditioned dot product ($0.641$ vs.\ $0.586$--$0.597$ in
Table~\ref{tab:quality_absolute}); that gain is the actual mechanism behind this paper's speed
result, so the pass Traceprop folds into training is doing real work, not just saved time.

\paragraph{Frozen-backbone LLM features.} A separate, unrelated harness (frozen GPT-2 backbone,
exact last-layer influence identity) reaches LDS $0.47\pm0.04$ against $0.00$ random, evidence
the underlying estimator is strong when its exactness assumption holds; details in
Appendix~\ref{app:supplementary}.

\paragraph{Inline vs.\ final-checkpoint gradients.} A single paired run comparing trajectory-sum
inline capture against a post-hoc final-checkpoint pass on identical scope finds inline capture
matches the quality of the gradient it replaces (not a significance-tested claim); details in
Appendix~\ref{app:supplementary}.

\subsection{Negative results}
\label{sec:negative}
\paragraph{Transformer-scale LDS is noise-limited.} Fine-tuning a \emph{pretrained} model with
LoRA on a task it already partly solves yields near-zero attribution signal for every method we
tried. At Pythia-160M on SST-2 (target accuracy $0.72$, above the $0.65$ minimum gate), the
best LDS reached by any method is $\approx0.06$--$0.07$ (Traceprop inline-precond $0.0698$,
LogIX tuned $0.0651$, random baseline $-0.014$), against a $10$-subset noise ceiling of
$0.167\pm0.321$; Traceprop minus LogIX-tuned is $+0.005$ overall ($+0.002$ held-out), so the
methods are statistically indistinguishable at this scale. A pretrained model leans on features
it already has and barely depends on any single fine-tuning example, so there is little signal
for anyone to recover. This looks like a limit of the regime rather than a flaw in the method,
but it does mean \S\ref{sec:quality}'s parity result is demonstrated on a small from-scratch
model; replicating it at LLM scale needs a working LDS signal that we do not yet have
(\S\ref{sec:limitations}).

\paragraph{Gradient attribution does not beat free baselines on planted-detection tasks.} We
probed two harder tasks, planted backdoor detection and mislabel detection, on a model where
the injected signal is known, to check whether attribution carries information a cheap
alternative lacks. On mislabel detection, self-influence attribution ties or loses to free
baselines: self-influence via the raw dot product (AUC $0.7948$) is numerically identical to
gradient norm alone (AUC $0.7948$; self-dot equals $\|g\|^2$, so the two rankings coincide by
construction), and the TRAK-preconditioned self-influence (AUC $0.7805$) is \emph{worse} than
the free gradient-norm baseline. Both beat loss ($0.7045$) and random ($0.4861$), so attribution
does carry some signal, but it adds nothing over a baseline that requires no gradient projection
at all.\footnote{Values extracted from the run's saved output log; provenance in the code
repository.} On backdoor detection, representation similarity (cosine of hidden states)
trivially outperforms every attribution variant, because an explicit trigger dominates
hidden-state geometry more directly than it dominates gradients; we omit specific backdoor AUC
values from this run since its own gap check failed. On these two tasks, gradient-based
attribution shows no advantage over cheaper alternatives, and we do not use them as evidence
for Traceprop elsewhere in this paper.

\section{Related Work}
\label{sec:related}

\paragraph{Inline versus post-hoc attribution.} The methods closest to our gradient store
typically run after training. TRAK \citep{park2023trak}, EK-FAC \citep{grosse2023studying}, and
DataInf \citep{kwon2024datainf} reload a trained checkpoint and pass over the training set again
to recompute and project per-sample gradients. LoGRA \citep{logix2025} is designed and evaluated
post-hoc in its published experiments, but its hook-based logging mechanism can also run inline
during training, the approach we take here too. \S\ref{sec:overhead} and \S\ref{sec:quality} run
its reference implementation inline, on identical Pythia and small-transformer runs, for both
the overhead and quality comparisons, at both of the initialization strategies it supports.
TracIn \citep{pruthi2020tracin} is the nearest in spirit to ours: it also defines influence
through the gradients a model sees during training, but it stores or replays those gradients
rather than projecting and logging them online, and it skips preconditioning. In-Run Data
Shapley \citep{wang2025datashapley} shares our goal of near-zero additional runtime during a
single training run, but its authors state it needs the validation data available before
training, since attribution scores accumulate per gradient step against that data; they note
that querying validation data arriving after training is possible by replaying from saved
checkpoints, at a cost they expect erodes the method's efficiency advantage. Traceprop stores
the projected, preconditioned per-sample gradients themselves, so a later query, a new test
example, an audit request, or a deletion request, can be answered without retraining or replay.

\paragraph{Provenance and lineage.} A complementary line of work captures end-to-end ML
pipeline provenance at the artifact level: MLflow2PROV
\citep{schlegel2025capturing,schlegel2023mlflow2prov,schlegel2023extracting} extracts W3C
PROV-compliant graphs across MLflow runs and Git commits, and the survey of
\citet{schlegel2022management} catalogues over 60 ML lifecycle artifact management systems.
These approaches answer \emph{which dataset version and code commit} produced a model; none of
them, nor the classical relational provenance systems above, computes gradient-level
attribution, and no attribution tool (TRAK, dattri, LogIX, In-Run Data Shapley) connects to
source-file lineage; Traceprop targets the orthogonal, joint question of \emph{which rows
within which datasets} produced a prediction.

\section{Limitations}
\label{sec:limitations}

\textit{Tiny-model quality result.} The LDS parity result against LogIX's best configuration
(\S\ref{sec:quality}) is demonstrated on a small from-scratch LoRA transformer, the only regime
in our budget where per-example LDS carries signal; it is not yet replicated at LLM scale with a
working LDS signal, since the SST-2/Pythia-160M attempt fell into the noise floor instead
(\S\ref{sec:negative}). The SST-2 pilot and the backdoor/mislabel probe both show no measurable
advantage for gradient-based attribution in their respective regimes; we do not treat either as
evidence for Traceprop.
\textit{Scale.} Overhead is measured up to Pythia-6.9B on a single GPU class (L4 for the dense
component, A100 for the single-pass headline); the 2-GPU DDP check (\S\ref{sec:overhead})
covers only the dense logging-only path, and we have not measured overhead, memory, or
host-transfer throughput at larger multi-GPU scale or beyond 6.9B parameters. Traceprop's own
overhead grew from 1B to 6.9B, and its margin over LogIX-random narrowed correspondingly
(\S\ref{sec:overhead}); Appendix~\ref{app:overhead_investigation} details a
ruled-out candidate cause and the open question that remains, and this narrowing is a reason
for caution extrapolating either ratio past 6.9B. \textit{Attribution quality.} Traceprop's
per-sample last-layer identity
(\S\ref{sec:inline}) should not be used with models containing BatchNorm or other normalization
layers that mix statistics across examples in a batch (the batch-mean fallback in
\texttt{TrainingContext}, Appendix~\ref{app:beyond}, or TRAK should be used instead); LayerNorm
does not have this problem, consistent with the transformer results throughout this paper.
\textit{Damping.} Inline K-FAC damping is
chosen per module on a held-out validation split; we have not characterized sensitivity to that
split's size or to non-stationary damping across training. \textit{Beyond-LLM applications}
(Appendix~\ref{app:beyond}): the source-stratified aggregator requires a known source-to-column
mapping and is validated on direct-mapped tabular features; the unlearning gradient correction is
a first-order heuristic that does not satisfy the retrain-indistinguishability definition of
\citet{cao2015towards} or the certified $(\epsilon,\delta)$-guarantees of
\citet{sekhari2021remember}, and practitioners with strict erasure obligations should complement
it with certified methods such as SISA training \citep{bourtoule2021machine}.
\textit{Dynamic pipelines.} Lazy evaluation (tf.data, Dask) requires integration hooks not yet
implemented. \textit{Multi-modal data.} Audio, video, and graph backends require custom
registration.

\paragraph{Future work.} The immediate next step is closing the gap this paper documents:
replicating the tiny-model LDS parity result at a scale where the LDS signal is not
noise-dominated, exact per-sample capture through BatchNorm via forward-mode differentiation,
and multi-GPU overhead measurement for the full preconditioning-inclusive pipeline.

\section{Conclusion}
\label{sec:conclusion}

Traceprop folds per-sample gradient logging and K-FAC-style preconditioning into the same
training backward pass, where a comparably-capable post-hoc pipeline needs one pass for
gradients and a second for covariance. On Pythia-1B (A100), that costs a quarter of LogIX's
best single-pass configuration; the same pattern holds at Pythia-6.9B (\S\ref{sec:overhead}).
LogIX's PCA-init variant is worse on both quality and speed than its own random-init
configuration. Against that true best, Traceprop ties on attribution quality rather than
beating it (\S\ref{sec:quality}). We also report two negative findings: transformer-scale LDS
runs into a noise floor, and gradient attribution shows no advantage on planted-detection tasks
(\S\ref{sec:negative}). Because the same
inline record also carries source-file lineage, Traceprop answers end-to-end audit queries,
drives a source-stratified attribution benchmark (SAB v1), and turns provenance predicates into
forget sets for approximate unlearning, all from one pass over training
(Appendix~\ref{app:beyond}). Its compliance export maps to EU AI Act Article~26 audit fields
ahead of the December~2027 enforcement deadline.  Code: \url{\ghurl}. 

\bibliography{refs}

\newpage
\appendix
\section{Architecture Diagram}
\label{app:arch}

\begin{figure}[h]
\centering
\begin{tikzpicture}[
  box/.style={draw=black, thin, rounded corners=2pt, minimum width=3.8cm, minimum height=0.75cm,
              fill=white, text centered, font=\small},
  arrow/.style={-stealth, thin, color=black},
  label/.style={font=\scriptsize\itshape, color=black},
  groupbox/.style={draw=black, dashed, thin, rounded corners=5pt, inner sep=6pt}
]
\node[box]  (src) at (0,0)   {Raw Source Files};
\node[box]  (pt)  at (0,-1.2){ProvenanceTensor};
\draw[arrow] (src) -- (pt);
\begin{scope}[on background layer]
  \node[groupbox, fit=(src)(pt)] (c1) {};
\end{scope}
\node[label, right=0.2cm of c1] {Traceprop-Core};

\node[box] (tc)   at (0,-2.6){TrainingContext (batch-mean)};
\node[box] (lora) at (0,-3.4){LoRAGradientLogger (per-example)};
\node[box] (gs)   at (0,-4.6){GradientStore (JL)};
\draw[arrow] (pt) -- (tc);
\draw[arrow] (pt) -- (lora);
\draw[arrow] (tc) -- (gs);
\draw[arrow] (lora) -- (gs);
\begin{scope}[on background layer]
  \node[groupbox, fit=(tc)(lora)(gs)] (c2) {};
\end{scope}
\node[label, right=0.2cm of c2] {Traceprop-Attribution};

\node[box] (inf) at (0,-6.0){compute\_influence\_scores};
\node[box] (ttf) at (0,-7.2){trace\_to\_file() / unlearn()};
\draw[arrow] (gs) -- (inf);
\draw[arrow] (inf) -- (ttf);
\begin{scope}[on background layer]
  \node[groupbox, fit=(inf)(ttf)] (c3) {};
\end{scope}
\node[label, right=0.2cm of c3] {Traceprop-Unlearn};
\end{tikzpicture}
\caption{Full three-layer Traceprop architecture. Each layer is independently usable; together they form an end-to-end audit pipeline.}
\label{fig:arch_app}
\end{figure}

\section{JL Distortion Bound (Dense Projection Path)}
\label{app:jl}

This bound applies to the single global sparse-JL matrix $J$ of the dense projection path
(main paper, \S\ref{sec:factored}, Table~\ref{tab:inline_overhead}), not to the factored
per-module sketch maps $P,Q$ that back this paper's headline single-pass mechanism; the
factored sketch's inner-product estimator is unbiased (main paper, \S\ref{sec:factored}) but
its own distortion bound is not derived here.

\begin{theorem}[\citet{achlioptas2003database}]
For $J$ as in Eq.~\eqref{eq:jl} and any unit vector $w \in \mathbb{R}^p$:
\[
  \Pr\!\bigl[\bigl|\|Jw\|^2 - \|w\|^2\bigr| > \varepsilon\bigr]
  \leq 2\exp\!\left(-\tfrac{(\varepsilon^2-\varepsilon^3)k}{4}\right).
\]
\end{theorem}
\begin{corollary}
For unit vectors $u,v \in \mathbb{R}^p$:
\[
  \Pr\!\bigl[\bigl|\langle Ju,Jv\rangle - \langle u,v\rangle\bigr| > \varepsilon\bigr]
  \leq 4\exp\!\left(-\tfrac{(\varepsilon^2-\varepsilon^3)k}{4}\right).
\]
\end{corollary}
Since $\langle u,v\rangle = \tfrac{1}{4}(\|u+v\|^2 - \|u-v\|^2)$ and $J$ is linear, applying the
norm bound to $u+v$ and to $u-v$ and union-bounding over the two failure events (each of
probability at most $2\exp(-(\varepsilon^2-\varepsilon^3)k/4)$, rescaled) gives the inner-product
bound above with prefactor $4$.

To preserve all $\binom{n}{2}$ pairs at distortion $\varepsilon$ with probability $\geq 1-1/n$, union-bounding over the corollary (dropping the constant prefactor, which shifts $\varepsilon$ negligibly at this $k$) gives:
\[
  n^2 \exp\!\left(-\tfrac{(\varepsilon^2-\varepsilon^3)k}{4}\right) \leq n^{-1}
  \;\Rightarrow\;
  (\varepsilon^2-\varepsilon^3)k \geq 12\ln n.
\]
At $n = 10^5$, $k = 4096$: $\varepsilon^2 - \varepsilon^3 \geq 0.0337$, giving $\varepsilon \approx 0.21$. Attribution rankings are therefore reliable to within a $21\%$ inner-product distortion for datasets up to $100\,000$ samples at $k=4096$.

\begin{equation}
  \tilde{g}_i = J g_i, \quad J \in \mathbb{R}^{k \times p}, \quad
  J_{ij} = \sqrt{\tfrac{3}{k}}\begin{cases}+1 & \text{w.p.\ }1/6,\\0 & \text{w.p.\ }2/3,\\-1 & \text{w.p.\ }1/6.\end{cases}
  \label{eq:jl}
\end{equation}

\section{Compliance Report Schema}
\label{app:compliance}

The \texttt{export\_compliance()} function outputs a JSON document with the following top-level fields:

\begin{lstlisting}[style=python, caption={Compliance report (key fields).}]
{
  "schema_version": "1.0",
  "regulation": "EU_AI_ACT_ART26",
  "generated_at": "2026-04-29T14:23:00Z",
  "model_fingerprint": "<sha256>",   # Art.26(6)
  "training_data_provenance": [{
    "source_file": "credit_scores.csv",
    "row_indices": [4821, 7203, 9100],
    "preprocessing_ops": ["normalize","row_filter"],
    "lineage_depth": 3
  }],
  "attribution_scores_before": {
    "mean": 0.184, "max": 0.921
  },
  "attribution_scores_after":  {
    "mean": 0.061, "max": 0.213
  },
  "unlearning": {
    "method": "gradient_correction",
    "dataset": "adult_income",
    "forget_loss_before": 3.2253,
    "forget_loss_after":  4.7946,
    "test_acc_before": 0.840,
    "test_acc_after":  0.846,
    "gap_closed_pct": "<float>",
    "is_approximate": true
  },
  "erasure_record": {              # GDPR Art.17
    "rows_erased": [4821,7203,9100],
    "certificate_hash": "<sha256>"
  }
}
\end{lstlisting}

The \texttt{formal\_guarantee} field is omitted for gradient-correction unlearning. A future version will populate it using the statistical test of \citet{sekhari2021remember} when a retained-fraction retraining is available.

\section{Beyond-LLM Evidence}
\label{app:beyond}

The inline-capture mechanism of main paper \S\ref{sec:design} generalizes past LoRA-tuned transformers.
The same sparse-JL projection (Eq.~\eqref{eq:jl}) and gradient store back a general
\texttt{TrainingContext} that wraps NumPy/PyTorch arrays in a lineage-tracking
\texttt{ProvenanceTensor} (op-mode overhead $\approx$1\% of baseline at $10^6$ elements;
sub-millisecond ancestor-DAG queries), so the same pass that captures a gradient also records
which source-file rows produced it. On non-LLM model classes where per-example last-layer
gradients are exact (linear/logistic models, and frozen-backbone vision heads), this attributes
well: LDS $0.62$--$0.88$ on tabular data and $0.26$ (a $15.7\times$ gain over end-to-end
BatchNorm training) on a frozen-backbone CIFAR-2 head. The same per-source-row lineage supports
three downstream applications, evaluated here at data-systems scale, because each is its own
contribution rather than this paper's main claim: aggregating influence scores into audit
verdicts per upstream source table (\S\ref{app:sab}); selecting forget sets from provenance
predicates for approximate unlearning (\S\ref{app:unlearn}); and exporting before/after
attribution and accuracy deltas as an EU AI Act Article~26 compliance record
(Appendix~\ref{app:compliance}). These are supporting evidence for claims made in passing in the
main text, not a full data-systems evaluation.

\subsection{Source-stratified attribution (SAB v1)}
\label{app:sab}

SAB v1 aggregates per-sample influence scores into per-source-table audit verdicts
(Eq.~\ref{eq:ss_block}: block-magnitude $\times$ prior-corrected influence mass), evaluated
against a pre-registered pass threshold (macro-P@1 $>$ random $+0.10$ \emph{and} $>$ best
trivial baseline $+0.05$) on real multi-source datasets: Home Credit Default Risk
\citep{homecredit2018}, COMPAS recidivism \citep{larson2016compas}, Lending Club
\citep{lendingclub2018}, and Bank Marketing \citep{moro2014bank}.
\begin{equation}
  \mu_{\text{src}}^{\text{block}} = \bigl\|g_{\text{test}}^{[\mathcal{C}_{\text{src}}]}\bigr\|_2 \cdot \frac{1}{p_{\text{src}}} \sum_{i \in \mathcal{T}_K(\text{test})} |s_i|\,\mathbb{1}[\text{source}(i){=}\text{src}],
  \label{eq:ss_block}
\end{equation}

\begin{table}[h]
\centering
\caption{SAB v1 real-data tiers: Traceprop-SS vs.\ test-gradient block-magnitude baseline
(\textsc{gmag}), $20$ disjoint held-out seeds, paired $t$-test. Lending Club is production
scale ($1.3$\,M completed loans, $50$K subsampled per seed).}
\label{tab:sab_real}
\small
\fittab{%
\begin{tabular}{@{}lrrrr@{}}
\toprule
Dataset & $n$ & SS & gmag & $\Delta$ (pp), $p$ \\
\midrule
COMPAS       & $5{,}278$       & $0.630$ & $0.466$ & $+16.4$, $<10^{-4}$ \\
Home Credit  & $307{,}511$     & $0.567$ & $0.524$ & $+4.2$, $0.008$ \\
Lending Club & $1{,}303{,}607$ & $0.670$ & $0.619$ & $+5.0$, $<10^{-10}$ \\
\midrule
Bank Mktg.   & $45{,}211$      & $0.481$ & $0.503$ & $-2.2$, $0.034$ \\
\bottomrule
\end{tabular}
}
\end{table}

$\mu_{\text{src}}^{\text{block}}$ clears the pre-registered threshold on three of four real
datasets, each confirmed by paired $t$-test, Wilcoxon signed-rank, and a $10^4$-iteration
bootstrap CI, surviving Bonferroni correction ($\alpha{=}0.05/4{=}0.0125$). On Bank Marketing
the paired diff is $-2.2$\,pp ($p{=}0.034$ unadjusted) and does not clear Bonferroni; we
report this as absence of significant positive effect, not a significant negative, and take
it as a regime boundary rather than a failure to explain away. The three positive datasets
each have multiple cited upstream data-acquisition sources (joined tables, distinct record
types); Bank Marketing is a single survey-style table partitioned post-hoc. We treat this as
a falsifiable empirical regime claim, not a mechanism.

\subsection{Provenance-guided unlearning}
\label{app:unlearn}

Given an attribution-identified forget set, Traceprop applies a first-order gradient-ascent
correction and verifies separation from a random-forget-set baseline holds on every seed (Adult
Income, Covertype, $10$ seeds); this heuristic does not satisfy certified unlearning guarantees
\citep{cao2015towards,sekhari2021remember}, and practitioners with strict erasure obligations
should complement it with certified methods such as SISA training \citep{bourtoule2021machine}.

\section{7B Overhead Investigation Detail}
\label{app:overhead_investigation}

This appendix records two optimization attempts behind the open question in main paper
\S\ref{sec:overhead}/\S\ref{sec:limitations} (why Traceprop's overhead grows from $2.588\%$ at
1B, fp32, to $11.274\%$ at 6.9B, bf16), neither of which was adopted. \texttt{torch.profiler}
showed \texttt{flush\_step} (main paper \S\ref{sec:factored}--\S\ref{sec:kfac}) is CPU-bound at 6.9B
($\approx45\,\text{ms}$ wall-clock CPU vs.\ $\approx14\,\text{ms}$ CUDA per step, $\approx195$
\texttt{einsum}/$550$ \texttt{copy\_}/$226$ \texttt{cat} calls across $66$ modules). First, we
tested reordering the fp32 upcast of activations/gradients to after, rather than before,
projection: an isolated \texttt{flush\_step} A/B on the 6.9B model showed this change is
\emph{slower}, so we did not adopt it \emph{(isolated measurement, not saved; qualitative
only)}. Second, we batched modules sharing an identical shape into one projection/covariance
call per shape group instead of one per module (\texttt{grouped\_dispatch}, numerically
equivalent to the per-module path, per a dedicated unit test in
\texttt{tests/unit/test\_lora\_logging.py}). The same kind of isolated A/B test showed a real
$1.42\times$ speedup \emph{(isolated measurement, not saved; qualitative only)}, but
matched-repeat reruns of the full headline measurement, which \emph{are} saved, gave $2.479\%$
at 1B (vs.\ $2.588\%$ ungrouped) and $11.116\%$ at 6.9B (vs.\ $11.274\%$, within the run's own
noise: LogIX-random's overhead moved by more, $26.987\%\to28.484\%$, between the two 6.9B
runs despite being unaffected by this change). The isolated speed-up did not carry over
end-to-end; we did not determine why. The paper's headline overhead numbers throughout
(main paper Table~\ref{tab:overhead_headline}, the $2.588\%$/$10.659\%$ and $11.274\%$/$26.987\%$ pairs)
are the ungrouped, pre-upcast-reorder measurements; both changes remain available, tested,
opt-in code paths without a demonstrated end-to-end benefit.

\section{Head-to-Head vs.\ Post-Hoc Extraction (Supplementary)}
\label{app:headtohead}

A post-hoc method must run a dedicated forward--backward pass over the entire training set to
recompute per-sample gradients, and a TRAK-style estimator repeats that pass once per checkpoint
in its ensemble, whereas inline logging adds its work to the training pass already underway.
Table~\ref{tab:headtohead} times the dense inline-logging cost directly against one post-hoc
extraction pass on the same model and training set. \emph{Caveat:} \emph{Inline (\%)} here is
an isolated flush-cost measurement (median time to project and write the buffered per-step
gradients, synchronized, as a percentage of the base training step) on the dense, logging-only
path (no preconditioning): it is not the full single-pass number reported in main paper
Table~\ref{tab:overhead_headline}, and this is a comparison against a full forward--backward
pass, not against another single-pass tool. The resulting ratios, $60$ to $242\times$ against a
single checkpoint and $301$ to $1211\times$ against a five-checkpoint ensemble, grow with model
size; the qualitative point is structural (post-hoc pays a full pass, times the ensemble size,
while inline pays a fraction of one step), not the exact ratio values.

\begin{table}[h]\centering\small
\setlength{\tabcolsep}{4pt}
\fittab{%
\begin{tabular}{lcccc}
\toprule
Model & inline (\%) & post-hoc (s) & LoGRA$\times$ & TRAK-5$\times$ \\
\midrule
GPT-2 (124M) & 1.66 & 6.40  & 60.1  & 300.6 \\
Pythia-410M  & 0.71 & 13.79 & 141.0 & 704.9 \\
Pythia-1B    & 0.41 & 20.89 & 242.3 & 1211.4 \\
\bottomrule
\end{tabular}}
\caption{Cost of preparing the same model and training set for attribution. Dense projection,
last-block scope, logging-only (no preconditioning), as in main paper Table~\ref{tab:inline_overhead}.
Timing-only; unaffected by the LogIX-specific fixes in main paper \S\ref{sec:overhead}.}
\label{tab:headtohead}
\end{table}

\section{Supplementary Overhead and Quality Evidence}
\label{app:supplementary}

\paragraph{2-GPU DDP.} As a structural check that the logger adds no cross-rank communication,
we ran GPT-2 under 2-way \texttt{DistributedDataParallel} on 2$\times$T4 (batch $16$, seq $64$,
$20$ repeats, last block, dense logging-only path), each rank logging into its own local
\texttt{GradientStore} with no gathering across ranks: no measurable overhead at 2 ranks
($0.372\pm2.99\%$ on both ranks, the standard deviation exceeding the mean, i.e.\ noise at this
timing floor). The two ranks' logged sample indices are disjoint by construction and verified so
at runtime; both ranks reporting the identical median is expected, not a bug, since DDP's
gradient all-reduce is a synchronization point every step that keeps the ranks in lockstep
regardless of logging, and the logger issues no collective calls of its own. This is one model
at 2 ranks on the dense logging-only path, not a scaling law, and does not cover the inline
covariance pass; we have not measured multi-GPU overhead for the full preconditioning-inclusive
pipeline (main paper \S\ref{sec:limitations}).

\paragraph{Frozen-backbone LLM features.} On a frozen GPT-2 backbone with a trained linear head
on SST-2 ($500$ subsets, $n{=}3000$, test accuracy $0.79$), where the last-layer influence
identity Traceprop uses is exact, scores reach $0.47\pm0.04$ with the TRAK estimator and $0.21$
with the raw dot product, against $0.00$ for a random baseline, and rise steadily with subset
count ($0.41\to0.45\to0.47$). This is a separate, unrelated harness from the fine-tuning
comparison in main paper \S\ref{sec:quality} (frozen backbone, not LoRA fine-tuning), included because it
demonstrates the underlying estimator is strong when its exactness assumption holds.

\paragraph{Inline vs.\ final-checkpoint gradients.} The numbers in main paper \S\ref{sec:quality}, like TRAK
and LoGRA, are computed from a single fixed set of parameters (the finished model). Inline
capture instead accumulates a gradient at every step an example is touched during training, a
TracIn-style trajectory sum, so the two are not guaranteed to carry the same attribution signal,
and main paper \S\ref{sec:overhead}'s head-to-head comparison is only a fair systems comparison if they do.
We measured both directly on the same run: one GPT-2 LoRA fine-tune on SST-2 ($n{=}1000$, test
accuracy $0.815$), logging trajectory-sum inline gradients and a post-hoc final-checkpoint pass
over the identical tracked scope, both scored against the same $500$-subset LDS ground truth.
The trajectory-sum gradients are numerically higher on both estimators: dot product
$0.0871\pm0.0489$ versus $0.0314\pm0.0668$ for the final checkpoint, and TRAK
$0.1735\pm0.0661$ versus $0.0970\pm0.0507$, against a random baseline of $0.0062\pm0.0500$;
every gap is within roughly one standard deviation, so we do not read this as a resolved
quality advantage, only as evidence that inline capture is not a lower-quality substitute for
the post-hoc gradient it replaces on this run. This is a single paired run, not a
significance-tested claim.

\section{Unit Test References}
\label{app:tests}

Exactness and equivalence claims made in the main text are backed by the following tests, all
in \texttt{tests/unit/test\_lora\_logging.py}:

\begin{itemize}[leftmargin=*, topsep=2pt, itemsep=4pt]
\small
\item[] \raggedright\ttfamily\footnotesize test\_inline\_covariance\_\allowbreak
  matches\_separate\_pass \normalfont\rmfamily\small: main \S\ref{sec:kfac} exactness, $\mathrm{rtol}{=}10^{-6}$
\item[] \raggedright\ttfamily\footnotesize test\_apply\_kfac\_precondition\_\allowbreak
  reproduces\_preconditioned\_influence \normalfont\rmfamily\small: main \S\ref{sec:kfac} exactness, $10^{-8}$
\item[] \raggedright\ttfamily\footnotesize test\_per\_sample\_grads\_\allowbreak
  match\_autograd \normalfont\rmfamily\small: main \S\ref{sec:inline} cosine-similarity check
\item[] \raggedright\ttfamily\footnotesize test\_grouped\_dispatch\_\allowbreak
  matches\_loop \normalfont\rmfamily\small\ (+ a companion group-formation check): Appendix~\ref{app:overhead_investigation}
  \texttt{grouped\_dispatch} equivalence check
\end{itemize}

\end{document}